\newcommand{\softmax}{ \texttt{softmax}}
\newcommand{\loss} { \mathcal{L} }
\newcommand{\lpri} { \loss }

\newcommand{\ldata} {\mathcal{D}_{l}}

\newcommand{\D} {D}
\documentclass[11pt,a4paper]{article}
\usepackage[hyperref]{emnlp2020}
\usepackage{times}
\usepackage{latexsym}
\usepackage{ subcaption }

\usepackage{microtype}

\usepackage{amssymb,amsmath,latexsym,amsfonts,amsthm,cleveref}

\usepackage{CJKutf8}
\usepackage{amsmath,amsfonts}
\usepackage{booktabs}
\usepackage{multirow}
\usepackage{graphics}
\usepackage{booktabs}
\usepackage{graphicx}
\usepackage{subcaption}
\usepackage{microtype}

\aclfinalcopy 

\title{Dual-View Distilled BERT for Sentence Embedding}

\author{
  Xingyi Cheng \\
  Ant Group \\
  \texttt{fanyin.cxy@alibaba-inc.com} \\}

\date{}

\begin{document}
\maketitle

\begin{abstract}
Recently, BERT realized significant progress for sentence matching via word-level cross sentence attention.
However, the performance significantly drops when using siamese BERT-networks to derive two sentence embeddings, which fall short in capturing the global semantic since the word-level attention between two sentences is absent.
In this paper, we propose a Dual-view distilled BERT~(DvBERT) for sentence matching with sentence embeddings.
Our method deals with a sentence pair from two distinct views, i.e., Siamese View and Interaction View.
Siamese View is the backbone where we generate sentence embeddings. 
Interaction View integrates the cross sentence interaction as multiple teachers to boost the representation ability of sentence embeddings. 
Experiments on six STS tasks show that our method outperforms the state-of-the-art sentence embedding methods significantly. 

\end{abstract}

\maketitle

\section{Introduction}
Recent sentence representation models like BERT~\citep{DBLP:conf/naacl/DevlinCLT19} achieved state-of-the-art results on sentence-pair regression/classification tasks, such as question answering, natural language inference~(NLI)~\cite{DBLP:conf/emnlp/BowmanAPM15,DBLP:conf/naacl/WilliamsNB18}, and semantic textual similarity~(STS)~\cite{DBLP:conf/semeval/AgirreCDG12,agirre-etal-2013-sem,agirre-etal-2016-semeval,agirre-etal-2014-semeval,agirre-etal-2015-semeval}. 
However, it has a low computational efficiency when candidate sentence-pairs are not given ahead, leading to a massive computational overhead. 
For example, seeking the most relevant sentence-pair of a collection requires pairing all sentences.
The $O(n^2)$ computational complexity is an obstacle preventing many te{retrieval} applications from adopting the technology.

A standard method to reduce the computations is separately encoding each sentence into a vector representation and then compare any two of them by similarity distance.
However, in contrast to the standard BERT model, the performance of sentence matching is constrained.
For instance, SBERT~\citep{DBLP:conf/emnlp/ReimersG19} using the siamese BERT-networks that decreased the performance by 3-4 points evaluated by Spearman correlation~\cite{myers2004spearman} on STS-Benchmark~\cite{cer-etal-2017-semeval}, which implies room for improvement.
We argue that the siamese BERT-networks are limited to capture the full complexity of global semantic matching, neglecting the word-level interaction features across two sentences.
The feature has been proved vital for predicting matching degrees~\citep{DBLP:conf/coling/Lan018,xu2020symmetric}.

Motivated by these observations, we propose a Dual-view distilled BERT~(DvBERT) by incorporating the word-level interaction features into sentence embeddings while maintains the same efficiency as siamese BERT-networks. 
We take inspiration from Multi-view learning~\citep{DBLP:journals/corr/abs-1304-5634,DBLP:conf/emnlp/ClarkLML18} and train the sentence matching model from two views:
(1) Siamese View,
we start with the siamese BERT-networks as a backbone to derive sentence embeddings, to be able to capture semantics similarity efficiently by calculating distances on the two fixed-size vectors.
(2) Interaction View,
the standard pre-trained models with cross-sentence interactions are utilized, acting as multiple teachers that generate predictions about the training set provided to the siamese networks to learn. 
The association between the two views acts as a regularization term that trains a student with soft targets from the multiple teacher's output distributions, making the procedure similar to knowledge distillation~\cite{DBLP:journals/corr/HintonVD15}.
In contrast of other distilled versions of BERT~\cite{sanh2019distilbert,DBLP:conf/emnlp/SunCGL19}, our method aims to optimize sentence embedding representations with two heterogeneous networks, together with multi-task knowledge distillation~\cite{DBLP:journals/corr/abs-1904-09482}, neither distilling large models into a small model~\cite{DBLP:journals/corr/abs-1904-00796, DBLP:journals/corr/abs-1903-04190} nor born-again networks~\citep{DBLP:conf/icml/FurlanelloLTIA18,DBLP:conf/acl/ClarkLKML19}.
Besides, we compared the loss weighting and teacher annealing strategy~\cite{DBLP:conf/acl/ClarkLKML19} during the distillation process, suggesting that the latter was more efficient.
Experiments demonstrate that DvBERT can achieve superior performance than siamese BERT-networks on six STS datasets.



\section{Dual View Distilled BERT}
\begin{figure}
\centering 
\includegraphics[width=0.8\linewidth,height=0.7\linewidth]{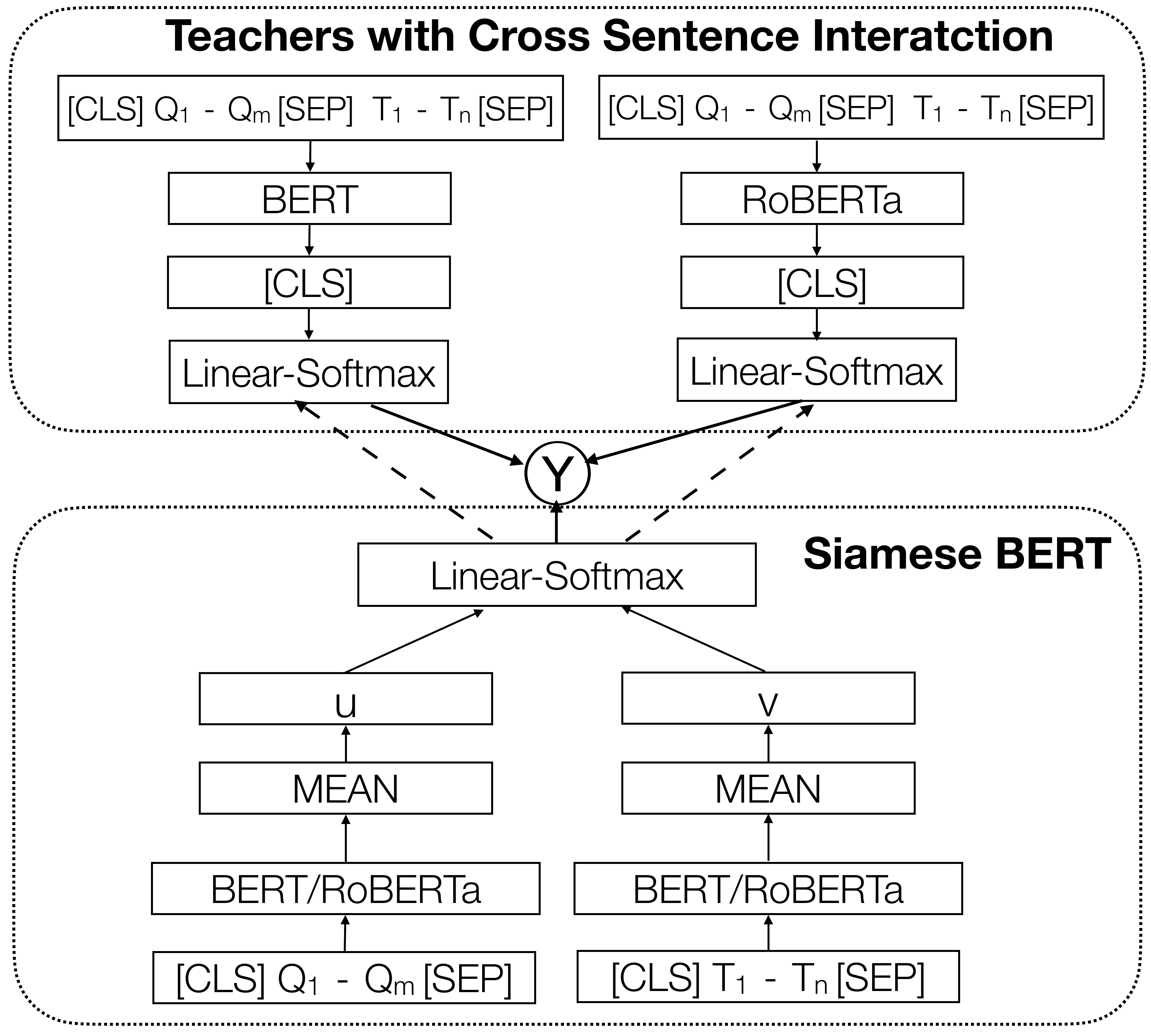}
\caption{Overview of Dual View Distilled BERT. Dash lines indicate distillation.}
\label{fig:framework}
\end{figure}

We first present DvBERT and describe how these views can be combined with multi-task knowledge distillation.


\begin{table*}[ht!]
\small
\centering
\resizebox{.7\textwidth}{!}{
\begin{tabular}{@{}l c c c c c c c c@{}}
\toprule
                        & STS12           & STS13  & STS14 & STS15 & STS16 & STS-B & Avg.\\ \midrule
BERT Avg. embedding             & 38.78 &57.98 &57.98 &63.15 &61.06 &46.35& 54.22              \\ 
BERT [CLS] embedding            & 20.16 &30.01 &20.09 &36.88 &38.08 &16.50&26.95             \\ \midrule
SBERT-base              &70.4 &71.77 &70.66&78.67&74.11	&76.28& 73.64         \\ 
SRoBERTa-base  &71.70 &	73.43 &	71.47 &\textbf{80.79} & 75.99 &77.02 & 75.06                      \\ 
DvBERT-base            &70.52	&73.17 &71.18&79.88	&75.08&77.96  & 74.63            \\ 
DvRoBERTa-base           &\textbf{72.42}	&\textbf{73.44}	&\textbf{72.21}	&80.43	&\textbf{76.52} &\textbf{78.32} &  \textbf{75.56}           \\
\midrule
SBERT-large             &71.68&72.79&72.20&80.32&76.45&78.00 &75.24               \\ 
SRoBERTa-large            & 72.14 &76.69 &74.12	 &79.81	&75.97	&78.60 & 76.22            \\
DvBERT-large            & 72.95&72.26&71.87&79.27&76.16&78.28 &75.13               \\ 
DvRoBERTa-large            & \textbf{74.99} &\textbf{76.16}	&\textbf{73.34} &\textbf{81.93}&\textbf{78.77}&\textbf{79.61} &\textbf{77.47}             \\ 
\bottomrule            
\end{tabular}
}
\caption{Spearman correlation of STS tasks without fine-tuning on task-specific data.}
\label{tab:unsup}
\end{table*}

\subsection{Siamese BERT-networks}
For a given dataset $\ldata$, Siamese BERT-networks aims to predict a label $y \in \mathbf{Y}$ by leveraging similarity measure between the sentence embeddings, where $\mathbf{Y} =\{entailment, contradiction, neutral\}$ in natural language inference.
For any sentence-pairs, the siamese BERT converts the two sentences into sequential vectors individually, and then pool these two vectors into two sentence embeddings $\mathbf{u}$ and $\mathbf{v}$. 
SBERT~\citep{DBLP:conf/emnlp/ReimersG19} compares different pooling strategies from multiple datasets, and gives the result that the \texttt{MEAN} strategy is significantly superior to \texttt{MAX} and \texttt{[CLS]} token strategy. Hereafter, the \texttt{MEAN} pooling is our default configuration. 
For classification tasks, such as NLI, we concatenate $\mathbf{u}$, $\mathbf{v}$, and $ |\mathbf{u} - \mathbf{v}|$ followed by a fully-connected layer, which projects the hidden size into a probability distribution.
\[
    p(y|\mathbf{u},\mathbf{v};\theta) = \softmax(W[\mathbf{u}, \mathbf{v}, |\mathbf{u} - \mathbf{v}|]),
\]
where $\theta$ represents all learnable parameters from BERT, shared for $\mathbf{u},\mathbf{v}$. And $W \in  \mathbb{R}^{3d \times n}$ is the parameter of the fully-connected layer. $d$ is the dimension of the sentence embeddings. We optimize the standard cross-entropy loss.

\subsection{Cross Sentence Interaction}
We use multiple teachers from different pre-trained models to introduce interaction matrices across words to enrich the word-level interactive features.
Each model first pre-trains with labeled data, then re-labeling the data and adds it to a new training set. 
Specifically, as illustrated in Fig~\ref{fig:framework}~(top), we concatenate the sentence-pair $Q=\{Q_i\}_{i=1,...,m}$ and $T=\{T_i\}_{i=1,...,N}$ into a text sequence $\texttt{[[CLS]} Q \texttt{[SEP]} T \texttt{[SEP]]}$. 
The $\texttt{[CLS]}$ token is regarded as an aggregated semantic gap of the input sentence-pair since it is used to predict whether a sentence-pair coherent or not during pre-training. Let $\mathbf{z}_{k}^{c}$ be the $\texttt{[CLS]}$ token from the $k$th pre-trained model, which followed a single fully-connected layer culminating in a softmax layer as our classifier:
\[
    q(y|\mathbf{z}_{k}^{c};\phi_k)= softmax(O\mathbf{z}_{k}^{c})),
\]
where $\phi_k$ and $O \in \mathbb{R}^{d \times n}$ are the model parameters. 
The siamese BERT learns from the hard targets as well as soft targets from the teachers.
Supposing that the $\phi_k$ and $O$ has been optimized by cross-entropy loss, DvBERT trains the siamese BERT by minimizing
\begin{align*}
      \lpri(\theta, W) = \sum_{k=1}^{K} \D(q(y|\mathbf{z}_{k}^{c};\phi_{k}), p(y|\mathbf{u},\mathbf{v};\theta)),
\label{soft_targets}
\end{align*}
where $\D$ is a distance function between probability distributions, here we use the KL-divergence. 
$K$ is the number of teachers.
We hold the teacher predictions $q(y|\mathbf{z}_{k}^{c};\phi)$ fixed when training the student. The BERT from the two views are not sharing since early experiments with sharing did not improve results.

\subsection{Teacher Annealing}
We leverage teacher annealing~\citep{DBLP:conf/acl/ClarkLKML19} strategy, which mixes the teacher prediction with the gold label during training. 
Teacher annealing progressively reduces the weight of soft targets as the training advances, making the student learning from the teacher to hard targets. 
This method ensures the student gets a rich training signal early in training but is not constrained to only simulating the teacher.
Specifically, summarizing the siamese BERT, and other $K$ BERT-related pre-trained model with cross sentence attention, the objective can be written as:
\begin{align*}
\resizebox{0.49\textwidth}{!}{
$\loss(\theta, W) = \sum_{k=1}^{K}D(\lambda y + (1 -\lambda) q(y|\mathbf{z}_{k}^{c};\phi_{k}), p(y|\mathbf{u},\mathbf{v};\theta))$
}
\end{align*}
where $\lambda$ is increase linearly from 0 to 1.
In the beginning, $\lambda=0$, which means the model is trained entirely based on the soft targets from teachers. 
As the model converges gradually,  the model learns from hard targets with more confidence.

\section{EXPERIMENTS}
In this section, we present our approach to NLI and STS datasets.
\subsection{Dataset}
The NLI dataset consists of SNLI~\citep{DBLP:conf/emnlp/BowmanAPM15} and MultiNLI~\citep{DBLP:conf/naacl/WilliamsNB18}, annotated with the labels contradiction, entailment, and neutral. 
STS~\citep{DBLP:conf/semeval/AgirreCDG12} assesses the matching degree to which two sentences are semantically equivalent to each other, which are human-annotated with a level of equivalence from 1 to 5. 
We follow the previous works~\citep{DBLP:conf/emnlp/ConneauKSBB17, DBLP:conf/emnlp/CerYKHLJCGYTSK18} to merge the training and test datasets in both NLI data as pre-training datasets of 940k sentence pairs. 
STS 2012-2016 datasets have no training data but 26k test data, so the datasets are used to evaluate the pre-trained DvBERT on NLI. 
STS-B is a collection of 8.6k sentence pairs and contains training, development, and test sets drawn from heterogeneous sources. 

\subsection{Training and Evaluation Settings}
We pre-train DvBERT with a 3-way softmax classifier for one epoch on NLI datasets. 
The batch size is set to 16, and the dropout rate is set to 0.1 for all modules. 
We use Adam optimizer~\citep{kingma2014adam} for model training.
We set the initial learning rate to 2e-5 with a decay ratio of 1.0, a linear learning rate warm-up over 10 percent.
For fune-tuning STS-B, we replace the $(u, v, |u-v|) $ to $cosine(u,v)$, and set the distance metric to the mean square error loss for regression training.
The epoch numbers were set to 4, and other hyperparameters keep the same as the NLI task setting. 
Basically, We keep hyper-parameters consistent with SBERT.
Our two default teachers are standard BERT, RoBERTa.  
We also evaluate the performance of DvRoBERTa by replacing the siamese BERT to RoBERTa.

\subsection{Unsupervised STS}
We apply STS 2012 - 2016 and STS-B test data to evaluate the performance without any task-specific training data.
We use the Spearman correlation between the cosine similarity of the sentence embeddings and the gold labels.
The results are reported in Table~\ref{tab:unsup}.
The first two lines show BERT without training on NLI get rather poor performance pooled by \texttt{MEAN} or \texttt{[CLS]} token.
Especially for \texttt{[CLS]} token, as it is mainly used to distinguish the segment-pair, whether coherence or not, there is a discrepancy in single sentence representations.
We evaluate our approach compared with SBERT~(SRoBERTa) on six STS datasets.
We can observe that the models with pre-training on NLI improve a large margin than those are not.
The dual-view method substantially impacts the performance of the two pre-trained models, obtaining 0.56\%-1.9\% improvement on average.

\begin{table}[ht!]
\small
\centering
\resizebox{.40\textwidth}{!}{
\begin{tabular}{@{}l c c c c @{}}
\toprule
 & \multicolumn{1}{c}{Base models} & \multicolumn{1}{c}{Large models} \\
\midrule
BERT-NLI          &  87.33 $\pm$ 0.23 & 89.09 $\pm$ 0.36 \\
RoBERTa-NLI     &  \textbf{89.77} $\pm$ 0.47 & \textbf{91.12} $\pm$ 0.17  \\
\midrule
SBERT        &  84.57   $\pm$ 0.2   & 84.72 $\pm$ 1.01  \\
SRoBERTa     &  84.89   $\pm$ 0.34  & 86.13 $\pm$ 0.94  \\
DvBERT      & 84.67 $\pm$ 0.23     & 85.31 $\pm$ 0.21           \\
DvRoBERTa       &  \textbf{85.31}  $\pm$ 0.37  &   \textbf{86.23} $\pm$ 0.67   \\
\midrule
SBERT-NLI        & 85.01   $\pm$ 0.17  & 85.91 $\pm$ 0.58   \\
SRoBERTa-NLI      &  85.40  $\pm$ 0.2  & 86.15 $\pm$ 0.35   \\
DvBERT-NLI        &  85.15 $\pm$ 0.24   &  86.21 $\pm$  0.13        \\
DvRoBERTa-NLI     & \textbf{86.05} $\pm$ 0.22   & \textbf{86.98} $\pm$ 0.46 \\
\bottomrule    

\end{tabular}
}
\caption{Spearman correlation of STS tasks. The average of 10 runs with different random seeds is reported. ``-NLI'' indicates the model is pre-trained on NLI data.}
\label{tab:sup}
\end{table}

\subsection{Fine-tuning on STS-B}
Since STS-B is a regression task, we adopt the cosine similarity followed mean square loss to take the place of both the fully-connected layer and cross-entropy loss from the NLI classification task.
The experiment was divided into three setups. (1) Standard BERT/RoBERTa pre-train on NLI, then fine-tune on STS-B; (2) DvBERT trains only on STSb; (3) DvBERT first trained on NLI for all teachers and the student, then trains on STS-B. 
The report gives the average Spearman correlation and its standard error after ten runs, as shown in table~\ref{tab:sup}.
The first two lines are standard BERT and RoBERTa models, which are used as our teacher models to capture the word-level attention to each other and significantly achieve the best results. 
We can observe that the pre-training on NLI consistently improves the performance for the shown models since NLI enhances the models towards language understanding.
The results demonstrate that DvBERT can improve generalization capability.

\subsection{Effect of Dual View Distillation}
\begin{figure}
\centering
\includegraphics[width=0.9\linewidth,height=0.45\linewidth]{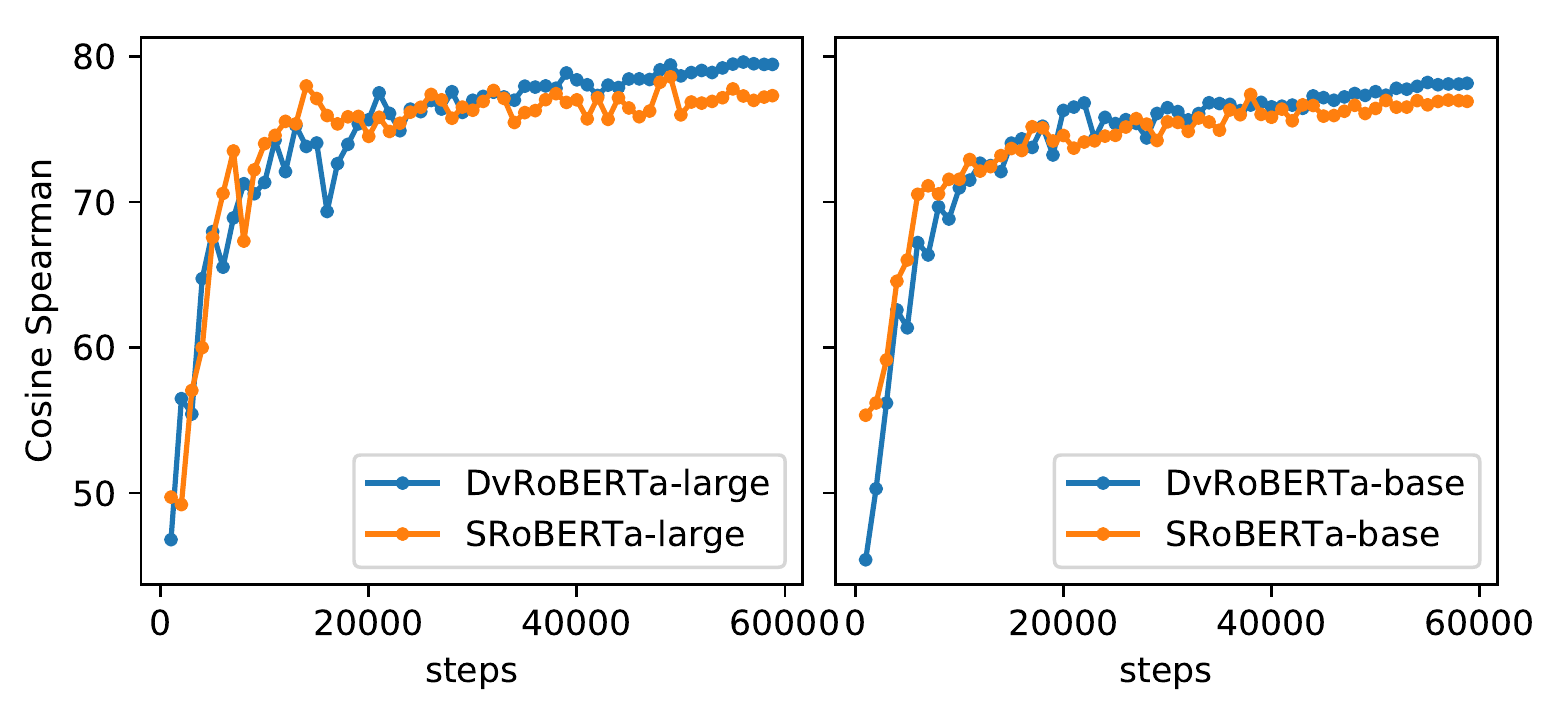}
\caption{Spearman correlation for SRoBERTa and DvRoBERTa.}
\label{fig:learning_curve}
\end{figure}

In order to observe how the dual-view method generalizes to STS set from the NLI training set, we plot the SRoBERTa vs. DvRoBERTa spearman correlation with every 1000 steps for one epoch.
The base model is configured with 12 layers, 12 self-attention heads, and the hidden size of 768 while the large model is set to 24 layers, 16 self-attention heads, and the hidden size of 1024. 
In Figure~\ref{fig:learning_curve}, we can see that both base and large models improve the ability to generalization.
We find the large model with dual-view achieves more benefits than the base model because of the large model of teachers.
Notably, DvRoBERTa has relatively poor performance in the early stage, as the student mainly learns from teachers, which are given higher weights to loss of soft targets.

\subsection{Effect of Teacher Annealing}
\label{sec:ta}
\begin{figure}[ht!]
\centering
\includegraphics[width=0.8\linewidth,height=0.5\linewidth]{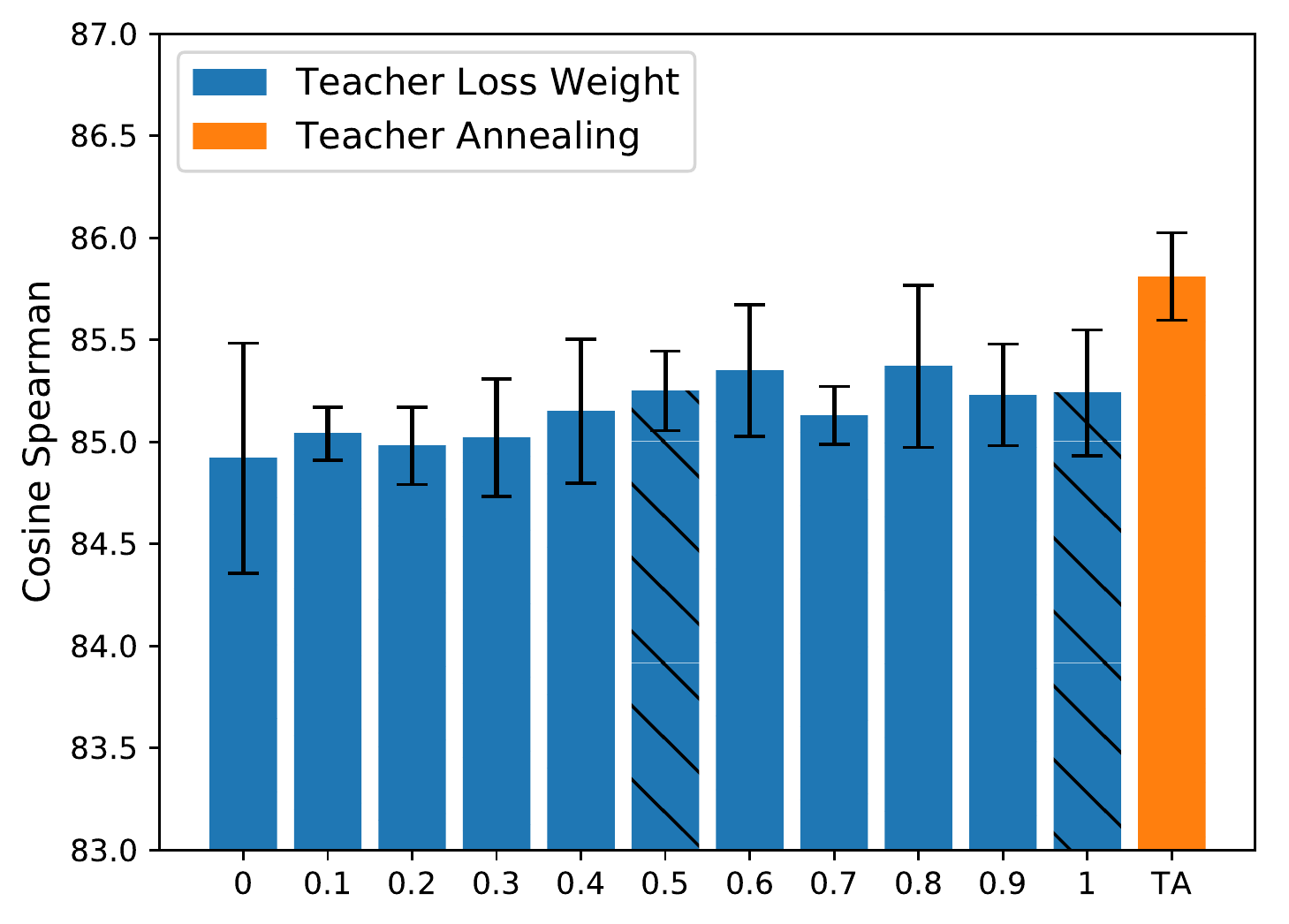}
\caption{Comparison of teacher loss Weighting and teacher annealing}
\label{fig:teacher_annealing}
\end{figure}

To verify the effect of teacher annealing strategy for DvBERT, we show the importance of teacher annealing vs. loss weighting strategy.
The loss weighting strategy combined the loss of hard targets and soft targets by weighted summation.
The hyper-parameter $\alpha$ from 0 to 1, it weights the loss of the soft target for optimization. 
As seen in Figure~\ref{fig:teacher_annealing}, the left eleven error bars show the cosine spearman correlation of DvRoBERTa-base with different the various $\alpha$. 
Using pure hard targets without teacher annealing~(i.e., $\alpha$ = 0) performs no better than weighted distillation.
It further illustrated the dual-views from sentence-pairs can boost the single view of siamese BERT. 
On the other hand, the teacher annealing strategy~(the right bar) shows a better correlation than the loss weighting strategy.

\section{CONCLUSIONS}
We proposed a dual-views approach that enhances sentence embeddings for matching, which adopt two heterogeneous networks to adapt to two views.
Specifically, it allows siamese BERT-networks to effectively leverage the cross sentence interaction models while keeping the efficiency of using sentence embedding in retrieval tasks.
The experiments on six STS datasets show that our models achieve consistent gains and outperform the performance of siamese BERT-networks.

\bibliography{anthology,emnlp2020}

\begin{thebibliography}{26}
\expandafter\ifx\csname natexlab\endcsname\relax\def\natexlab#1{#1}\fi

\bibitem[{Agirre et~al.(2015)Agirre, Banea, Cardie, Cer, Diab, Gonzalez-Agirre,
  Guo, Lopez-Gazpio, Maritxalar, Mihalcea, Rigau, Uria, and
  Wiebe}]{agirre-etal-2015-semeval}
Eneko Agirre, Carmen Banea, Claire Cardie, Daniel Cer, Mona Diab, Aitor
  Gonzalez-Agirre, Weiwei Guo, I{\~n}igo Lopez-Gazpio, Montse Maritxalar, Rada
  Mihalcea, German Rigau, Larraitz Uria, and Janyce Wiebe. 2015.
\newblock \href {https://doi.org/10.18653/v1/S15-2045} {{S}em{E}val-2015 task
  2: Semantic textual similarity, {E}nglish, {S}panish and pilot on
  interpretability}.
\newblock In \emph{Proceedings of the 9th International Workshop on Semantic
  Evaluation ({S}em{E}val 2015)}, pages 252--263, Denver, Colorado. Association
  for Computational Linguistics.

\bibitem[{Agirre et~al.(2014)Agirre, Banea, Cardie, Cer, Diab, Gonzalez-Agirre,
  Guo, Mihalcea, Rigau, and Wiebe}]{agirre-etal-2014-semeval}
Eneko Agirre, Carmen Banea, Claire Cardie, Daniel Cer, Mona Diab, Aitor
  Gonzalez-Agirre, Weiwei Guo, Rada Mihalcea, German Rigau, and Janyce Wiebe.
  2014.
\newblock \href {https://doi.org/10.3115/v1/S14-2010} {{S}em{E}val-2014 task
  10: Multilingual semantic textual similarity}.
\newblock In \emph{Proceedings of the 8th International Workshop on Semantic
  Evaluation ({S}em{E}val 2014)}, pages 81--91, Dublin, Ireland. Association
  for Computational Linguistics.

\bibitem[{Agirre et~al.(2016)Agirre, Banea, Cer, Diab, Gonzalez-Agirre,
  Mihalcea, Rigau, and Wiebe}]{agirre-etal-2016-semeval}
Eneko Agirre, Carmen Banea, Daniel Cer, Mona Diab, Aitor Gonzalez-Agirre, Rada
  Mihalcea, German Rigau, and Janyce Wiebe. 2016.
\newblock \href {https://doi.org/10.18653/v1/S16-1081} {{S}em{E}val-2016 task
  1: Semantic textual similarity, monolingual and cross-lingual evaluation}.
\newblock In \emph{Proceedings of the 10th International Workshop on Semantic
  Evaluation ({S}em{E}val-2016)}, pages 497--511, San Diego, California.
  Association for Computational Linguistics.

\bibitem[{Agirre et~al.(2013)Agirre, Cer, Diab, Gonzalez-Agirre, and
  Guo}]{agirre-etal-2013-sem}
Eneko Agirre, Daniel Cer, Mona Diab, Aitor Gonzalez-Agirre, and Weiwei Guo.
  2013.
\newblock \href {https://www.aclweb.org/anthology/S13-1004} {*{SEM} 2013 shared
  task: Semantic textual similarity}.
\newblock In \emph{Second Joint Conference on Lexical and Computational
  Semantics (*{SEM}), Volume 1: Proceedings of the Main Conference and the
  Shared Task: Semantic Textual Similarity}, pages 32--43, Atlanta, Georgia,
  USA. Association for Computational Linguistics.

\bibitem[{Agirre et~al.(2012)Agirre, Cer, Diab, and
  Gonzalez{-}Agirre}]{DBLP:conf/semeval/AgirreCDG12}
Eneko Agirre, Daniel~M. Cer, Mona~T. Diab, and Aitor Gonzalez{-}Agirre. 2012.
\newblock \href {https://www.aclweb.org/anthology/S12-1051/} {Semeval-2012 task
  6: {A} pilot on semantic textual similarity}.
\newblock In \emph{Proceedings of the 6th International Workshop on Semantic
  Evaluation, SemEval@NAACL-HLT 2012, Montr{\'{e}}al, Canada, June 7-8, 2012},
  pages 385--393.

\bibitem[{Bowman et~al.(2015)Bowman, Angeli, Potts, and
  Manning}]{DBLP:conf/emnlp/BowmanAPM15}
Samuel~R. Bowman, Gabor Angeli, Christopher Potts, and Christopher~D. Manning.
  2015.
\newblock \href {https://doi.org/10.18653/v1/d15-1075} {A large annotated
  corpus for learning natural language inference}.
\newblock In \emph{{EMNLP} 2015, Lisbon, Portugal, September 17-21, 2015},
  pages 632--642.

\bibitem[{Cer et~al.(2017)Cer, Diab, Agirre, Lopez-Gazpio, and
  Specia}]{cer-etal-2017-semeval}
Daniel Cer, Mona Diab, Eneko Agirre, I{\~n}igo Lopez-Gazpio, and Lucia Specia.
  2017.
\newblock \href {https://doi.org/10.18653/v1/S17-2001} {{S}em{E}val-2017 task
  1: Semantic textual similarity multilingual and crosslingual focused
  evaluation}.
\newblock In \emph{Proceedings of the 11th International Workshop on Semantic
  Evaluation ({S}em{E}val-2017)}, pages 1--14, Vancouver, Canada. Association
  for Computational Linguistics.

\bibitem[{Cer et~al.(2018)Cer, Yang, Kong, Hua, Limtiaco, John, Constant,
  Guajardo{-}Cespedes, Yuan, Tar, Strope, and
  Kurzweil}]{DBLP:conf/emnlp/CerYKHLJCGYTSK18}
Daniel Cer, Yinfei Yang, Sheng{-}yi Kong, Nan Hua, Nicole Limtiaco, Rhomni~St.
  John, Noah Constant, Mario Guajardo{-}Cespedes, Steve Yuan, Chris Tar, Brian
  Strope, and Ray Kurzweil. 2018.
\newblock \href {https://doi.org/10.18653/v1/d18-2029} {Universal sentence
  encoder for english}.
\newblock In \emph{{EMNLP} 2018: System Demonstrations, Brussels, Belgium,
  October 31 - November 4, 2018}, pages 169--174.

\bibitem[{Chatterjee(2019)}]{DBLP:journals/corr/abs-1904-00796}
Debajyoti Chatterjee. 2019.
\newblock \href {http://arxiv.org/abs/1904.00796} {Making neural machine
  reading comprehension faster}.
\newblock \emph{CoRR}, abs/1904.00796.

\bibitem[{Clark et~al.(2019)Clark, Luong, Khandelwal, Manning, and
  Le}]{DBLP:conf/acl/ClarkLKML19}
Kevin Clark, Minh{-}Thang Luong, Urvashi Khandelwal, Christopher~D. Manning,
  and Quoc~V. Le. 2019.
\newblock \href {https://doi.org/10.18653/v1/p19-1595} {Bam! born-again
  multi-task networks for natural language understanding}.
\newblock In \emph{Proceedings of the 57th Conference of the Association for
  Computational Linguistics, {ACL} 2019, Florence, Italy, July 28- August 2,
  2019, Volume 1: Long Papers}, pages 5931--5937. Association for Computational
  Linguistics.

\bibitem[{Clark et~al.(2018)Clark, Luong, Manning, and
  Le}]{DBLP:conf/emnlp/ClarkLML18}
Kevin Clark, Minh{-}Thang Luong, Christopher~D. Manning, and Quoc~V. Le. 2018.
\newblock \href {https://doi.org/10.18653/v1/d18-1217} {Semi-supervised
  sequence modeling with cross-view training}.
\newblock In \emph{Proceedings of the 2018 Conference on Empirical Methods in
  Natural Language Processing, Brussels, Belgium, October 31 - November 4,
  2018}, pages 1914--1925.

\bibitem[{Conneau et~al.(2017)Conneau, Kiela, Schwenk, Barrault, and
  Bordes}]{DBLP:conf/emnlp/ConneauKSBB17}
Alexis Conneau, Douwe Kiela, Holger Schwenk, Lo{\"{\i}}c Barrault, and Antoine
  Bordes. 2017.
\newblock \href {https://doi.org/10.18653/v1/d17-1070} {Supervised learning of
  universal sentence representations from natural language inference data}.
\newblock In \emph{{EMNLP} 2017, Copenhagen, Denmark, September 9-11, 2017},
  pages 670--680.

\bibitem[{Devlin et~al.(2019)Devlin, Chang, Lee, and
  Toutanova}]{DBLP:conf/naacl/DevlinCLT19}
Jacob Devlin, Ming{-}Wei Chang, Kenton Lee, and Kristina Toutanova. 2019.
\newblock \href {https://doi.org/10.18653/v1/n19-1423} {{BERT:} pre-training of
  deep bidirectional transformers for language understanding}.
\newblock In \emph{{NAACL-HLT} 2019, Minneapolis, MN, USA, June 2-7, 2019,
  Volume 1 (Long and Short Papers)}, pages 4171--4186.

\bibitem[{Furlanello et~al.(2018)Furlanello, Lipton, Tschannen, Itti, and
  Anandkumar}]{DBLP:conf/icml/FurlanelloLTIA18}
Tommaso Furlanello, Zachary~Chase Lipton, Michael Tschannen, Laurent Itti, and
  Anima Anandkumar. 2018.
\newblock \href {http://proceedings.mlr.press/v80/furlanello18a.html}
  {Born-again neural networks}.
\newblock In \emph{Proceedings of the 35th International Conference on Machine
  Learning, {ICML} 2018, Stockholmsm{\"{a}}ssan, Stockholm, Sweden, July 10-15,
  2018}, volume~80 of \emph{Proceedings of Machine Learning Research}, pages
  1602--1611. {PMLR}.

\bibitem[{Hinton et~al.(2015)Hinton, Vinyals, and
  Dean}]{DBLP:journals/corr/HintonVD15}
Geoffrey~E. Hinton, Oriol Vinyals, and Jeffrey Dean. 2015.
\newblock \href {http://arxiv.org/abs/1503.02531} {Distilling the knowledge in
  a neural network}.
\newblock \emph{CoRR}, abs/1503.02531.

\bibitem[{Huang et~al.(2019)Huang, Cheng, Chen, Wang, and
  Chu}]{DBLP:journals/corr/abs-1903-04190}
Weipeng Huang, Xingyi Cheng, Kunlong Chen, Taifeng Wang, and Wei Chu. 2019.
\newblock \href {http://arxiv.org/abs/1903.04190} {Toward fast and accurate
  neural chinese word segmentation with multi-criteria learning}.
\newblock \emph{CoRR}, abs/1903.04190.

\bibitem[{Kingma and Ba(2014)}]{kingma2014adam}
Diederik~P Kingma and Jimmy Ba. 2014.
\newblock Adam: A method for stochastic optimization.
\newblock \emph{arXiv preprint arXiv:1412.6980}.

\bibitem[{Lan and Xu(2018)}]{DBLP:conf/coling/Lan018}
Wuwei Lan and Wei Xu. 2018.
\newblock \href {https://www.aclweb.org/anthology/C18-1328/} {Neural network
  models for paraphrase identification, semantic textual similarity, natural
  language inference, and question answering}.
\newblock In \emph{{COLING} 2018, Santa Fe, New Mexico, USA, August 20-26,
  2018}, pages 3890--3902.

\bibitem[{Liu et~al.(2019)Liu, He, Chen, and
  Gao}]{DBLP:journals/corr/abs-1904-09482}
Xiaodong Liu, Pengcheng He, Weizhu Chen, and Jianfeng Gao. 2019.
\newblock \href {http://arxiv.org/abs/1904.09482} {Improving multi-task deep
  neural networks via knowledge distillation for natural language
  understanding}.
\newblock \emph{CoRR}, abs/1904.09482.

\bibitem[{Myers and Sirois(2004)}]{myers2004spearman}
Leann Myers and Maria~J Sirois. 2004.
\newblock Spearman correlation coefficients, differences between.
\newblock \emph{Encyclopedia of statistical sciences}, 12.

\bibitem[{Reimers and Gurevych(2019)}]{DBLP:conf/emnlp/ReimersG19}
Nils Reimers and Iryna Gurevych. 2019.
\newblock \href {https://doi.org/10.18653/v1/D19-1410} {Sentence-bert: Sentence
  embeddings using siamese bert-networks}.
\newblock In \emph{{EMNLP-IJCNLP} 2019, Hong Kong, China, November 3-7, 2019},
  pages 3980--3990.

\bibitem[{Sanh et~al.(2019)Sanh, Debut, Chaumond, and
  Wolf}]{sanh2019distilbert}
Victor Sanh, Lysandre Debut, Julien Chaumond, and Thomas Wolf. 2019.
\newblock Distilbert, a distilled version of bert: smaller, faster, cheaper and
  lighter.
\newblock \emph{arXiv preprint arXiv:1910.01108}.

\bibitem[{Sun et~al.(2019)Sun, Cheng, Gan, and Liu}]{DBLP:conf/emnlp/SunCGL19}
Siqi Sun, Yu~Cheng, Zhe Gan, and Jingjing Liu. 2019.
\newblock \href {https://doi.org/10.18653/v1/D19-1441} {Patient knowledge
  distillation for {BERT} model compression}.
\newblock In \emph{Proceedings of the 2019 Conference on Empirical Methods in
  Natural Language Processing and the 9th International Joint Conference on
  Natural Language Processing, {EMNLP-IJCNLP} 2019, Hong Kong, China, November
  3-7, 2019}, pages 4322--4331. Association for Computational Linguistics.

\bibitem[{Williams et~al.(2018)Williams, Nangia, and
  Bowman}]{DBLP:conf/naacl/WilliamsNB18}
Adina Williams, Nikita Nangia, and Samuel~R. Bowman. 2018.
\newblock \href {https://doi.org/10.18653/v1/n18-1101} {A broad-coverage
  challenge corpus for sentence understanding through inference}.
\newblock In \emph{{NAACL-HLT} 2018, New Orleans, Louisiana, USA, June 1-6,
  2018, Volume 1 (Long Papers)}, pages 1112--1122.

\bibitem[{Xu et~al.(2013)Xu, Tao, and Xu}]{DBLP:journals/corr/abs-1304-5634}
Chang Xu, Dacheng Tao, and Chao Xu. 2013.
\newblock \href {http://arxiv.org/abs/1304.5634} {A survey on multi-view
  learning}.
\newblock \emph{CoRR}, abs/1304.5634.

\bibitem[{Xu et~al.(2020)Xu, Cheng, Chen, and Wang}]{xu2020symmetric}
Weidi Xu, Xingyi Cheng, Kunlong Chen, and Taifeng Wang. 2020.
\newblock Symmetric regularization based bert for pair-wise semantic reasoning.
\newblock In \emph{Proceedings of the 43rd International ACM SIGIR Conference
  on Research and Development in Information Retrieval}, pages 1901--1904.

\end{thebibliography}
\bibliographystyle{acl_natbib}

\end{document}